\documentclass[sigconf]{acmart}
\AtBeginDocument{%
  }

\usepackage{algorithm}
\usepackage{algorithmic}
\usepackage{multirow}
\usepackage{colortbl}

\usepackage{bbding}
\usepackage{pifont}
\usepackage{makecell}
\usepackage{subcaption}

\usepackage{graphicx}
\usepackage{subcaption}
\usepackage{pifont}
\usepackage{bbding}
\usepackage{graphicx}
\usepackage{array}
\usepackage{relsize}

\setcopyright{acmlicensed}
\copyrightyear{2025}
\acmYear{2025}
\acmConference[MM '25] {Proceedings of the 33rd ACM International Conference on Multimedia}{October 27--31, 2025}{Dublin, Ireland.}
\acmBooktitle{Proceedings of the 33rd ACM International Conference on Multimedia (MM '25), October 27--31, 2025, Dublin, Ireland}
\acmISBN{979-8-4007-2035-2/2025/10}
\acmDOI{10.1145/3746027.3754733} % https://doi.org/10.1145/3746027.3754733

\begin{document}

\title{Enhancing Pseudo-Boxes via Data-Level LiDAR-Camera Fusion for Unsupervised 3D Object Detection}

\author{Mingqian Ji}
\affiliation{
\institution{PCA Lab, School of Computer Science and Engineering, Nanjing University of Science and Technology}
\city{Nanjing}
\state{Jiangsu}
\country{China}
}
\email{mingqianji@njust.edu.cn}

\author{Jian Yang}
\affiliation{
\institution{PCA Lab, School of Computer Science and Engineering, Nanjing University of Science and Technology}
\city{Nanjing}
\state{Jiangsu}
\country{China}
}
\email{csjyang@njust.edu.cn}

\author{Shanshan Zhang}
\authornote{Corresponding author.}
\affiliation{
\institution{PCA Lab, School of Computer Science and Engineering, Nanjing University of Science and Technology}
\city{Nanjing}
\state{Jiangsu}
\country{China}
}
\email{shanshan.zhang@njust.edu.cn}

\renewcommand{\shortauthors}{Mingqian Ji, Jian Yang, \& Shanshan Zhang}

\begin{abstract}
   Existing LiDAR-based 3D object detectors typically rely on manually annotated labels for training to achieve good performance. However, obtaining high-quality 3D labels is time-consuming and labor-intensive. To address this issue, recent works explore unsupervised 3D object detection by introducing RGB images as an auxiliary modal to assist pseudo-box generation. However, these methods simply integrate pseudo-boxes generated by LiDAR point clouds and RGB images. Yet, such a label-level fusion strategy brings limited improvements to the quality of pseudo-boxes, as it overlooks the complementary nature in terms of LiDAR and RGB image data. To overcome the above limitations, we propose a novel data-level fusion framework that integrates RGB images and LiDAR data at an early stage. Specifically, we utilize vision foundation models for instance segmentation and depth estimation on images and introduce a bi-directional fusion method, where real points acquire category labels from the 2D space, while 2D pixels are projected onto 3D to enhance real point density. To mitigate noise from depth and segmentation estimations, we propose a local and global filtering method, which applies local radius filtering to suppress depth estimation errors and global statistical filtering to remove segmentation-induced outliers. Furthermore, we propose a data-level fusion based dynamic self-evolution strategy, which iteratively refines pseudo-boxes under a dense representation, significantly improving localization accuracy. Extensive experiments on the nuScenes dataset demonstrate that the detector trained by our method significantly outperforms that trained by previous state-of-the-art methods with 28.4$\%$ mAP on the nuScenes validation benchmark.
\end{abstract}







\maketitle

\section{Introduction}
 LiDAR-based 3D object detection plays a crucial role in autonomous driving. Recent LiDAR-based 3D object detectors \cite{yan2018second,lang2019pointpillars,shi2020pv,yin2021center} demonstrate strong performance on benchmark datasets \cite{geiger2012kitti,sun2020waymo,caesar2020nuscenes,xiao2021pandaset}. However, these methods rely heavily on manually annotated 3D bounding boxes, which are both time-consuming and labor-intensive to obtain. According to statistics \cite{wu2023efficient}, annotating a single 3D bounding box takes approximately five minutes for an experienced annotator, making large-scale annotation impractical. To get rid of expensive 3D annotations, we are highly motivated to investigate unsupervised learning approaches for 3D object detection.

    \begin{figure}[t!]
      \centering
      \includegraphics[width=1.0 \linewidth]{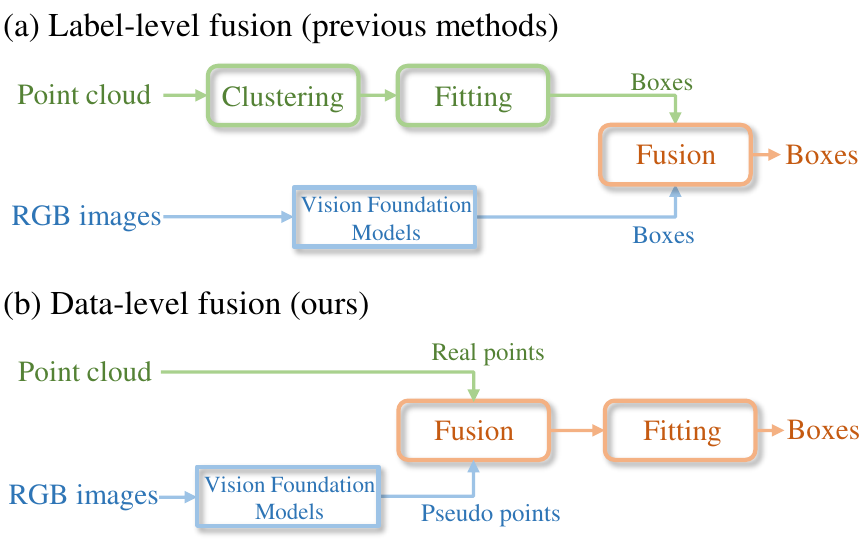}
      \caption{Comparison of previous multi-modal methods and ours. (a) Previous multi-modal methods combine information at the late stage through label-level fusion; (b) Our method integrates 2D and 3D data at the early stage using data-level fusion to generate pseudo-boxes.}
      \Description{}
      \label{intro}
    \end{figure}

 In recent years, unsupervised 3D object detection has received widespread attention. Early methods rely on clustering and fitting moving objects in LiDAR point clouds to generate initial pseudo-boxes, which are then iteratively refined through detector training. However, these approaches suffer from two major limitations: (1) they struggle to capture object categories, leading to category-agnostic pseudo-boxes that limit practical applicability; and (2) the quality of pseudo-boxes is constrained by the sparsity and incompleteness of point cloud data, resulting in a considerable gap between pseudo-boxes and ground truth annotations.
 
 To address these issues, recent multi-modal approaches incorporate image data as a complementary modality to enhance the quality of pseudo-boxes. As illustrated in Figure \ref{intro} (a), these approaches primarily leverage detection boxes produced from vision foundation models to modify the pseudo-boxes initially generated from the LiDAR point clouds. While this strategy mitigates some of the shortcomings of single-modal methods, this label-level refinement remains fundamentally reliant on point cloud structures, merely correcting existing boxes rather than fully utilizing the complementary strengths of image and LiDAR modalities. Consequently, pseudo-box quality remains suboptimal, as the method does not effectively compensate for missing information in sparse or occluded regions. Additionally, vision foundation models introduce noise due to their inherent estimation errors, further impacting the quality of pseudo-boxes. These limitations highlight the need for a more comprehensive fusion strategy that integrates image and point cloud data at the data level, rather than relying solely on label modification.
 
 To overcome these challenges, we propose DFU3D, a novel \textbf{data-level fusion} method for unsupervised 3D object detection. Unlike previous approaches that focus solely on post hoc pseudo-box refinement, DFU3D fully integrates image and point cloud information at the data level, enabling more comprehensive cross-modal fusion. Specifically, we employ vision foundation models to estimate the instance segmentation masks and depths of foreground objects in RGB images. Then, we propose \textbf{a bi-directional fusion method} that maps each real point back to the 2D RGB space to fetch its corresponding category label, and simultaneously maps each 2D RGB pixel to the 3D space to generate dense pseudo points, enriching the geometric and semantic representation. To further enhance fusion quality, we introduce \textbf{a local and global filtering method} to suppress noise introduced by errors in mask and depth estimation. We apply local radius filtering to remove noise points in the pseudo points caused by depth estimation errors; and use global statistical filtering to eliminate outliers introduced by segmentation errors, ensuring robust cross-modal consistency. With the cleaned dense points, we generate high-quality pseudo-boxes, mitigating the limitations of sparse and incomplete LiDAR data. To further improve the quality of pseudo-boxes, we employ \textbf{a data-level fusion based dynamic self-evolution strategy} to integrate the pseudo foreground points with the real points, and dynamically refine pseudo-boxes under these dense points, leading to more reliable pseudo-boxes. 
 
 By adopting early-stage data-level fusion rather than late-stage label refinement, our method better exploits cross-modal complementarities, resulting in more accurate pseudo-boxes and improved 3D object detection performance. Our main contributions are summarized as follows:
\begin{itemize}
    \item
    We propose a data-level fusion method for unsupervised 3D object detection, which includes a bi-directional fusion method and a local and global filtering method. The bi-directional fusion method allows real points to acquire class labels from 2D RGB space and projects 2D pixels into 3D to enhance the density of real points, and the local and global filtering method, consisting of local radius filtering and global statistical filtering, mitigates noise introduced by errors in mask and depth estimation, ensuring the high-quality pseudo-box generation.
    
    \item
    To further improve the quality of pseudo-boxes, we propose a data-level fusion based dynamic self-evolution paradigm that merges pseudo foreground points within updated boxes with the real points, and dynamically refines pseudo-boxes by locating local convergence points. This dynamic refinement under a dense representation significantly enhances pseudo-box accuracy.
    
    \item 
    Our method is evaluated on the nuScenes validation dataset, and shows to outperform previous state-of-the-art unsupervised methods in the way of providing more accurate detections.
\end{itemize}

    \begin{figure*}[t]
      \centering
       \includegraphics[width=1.0 \linewidth]{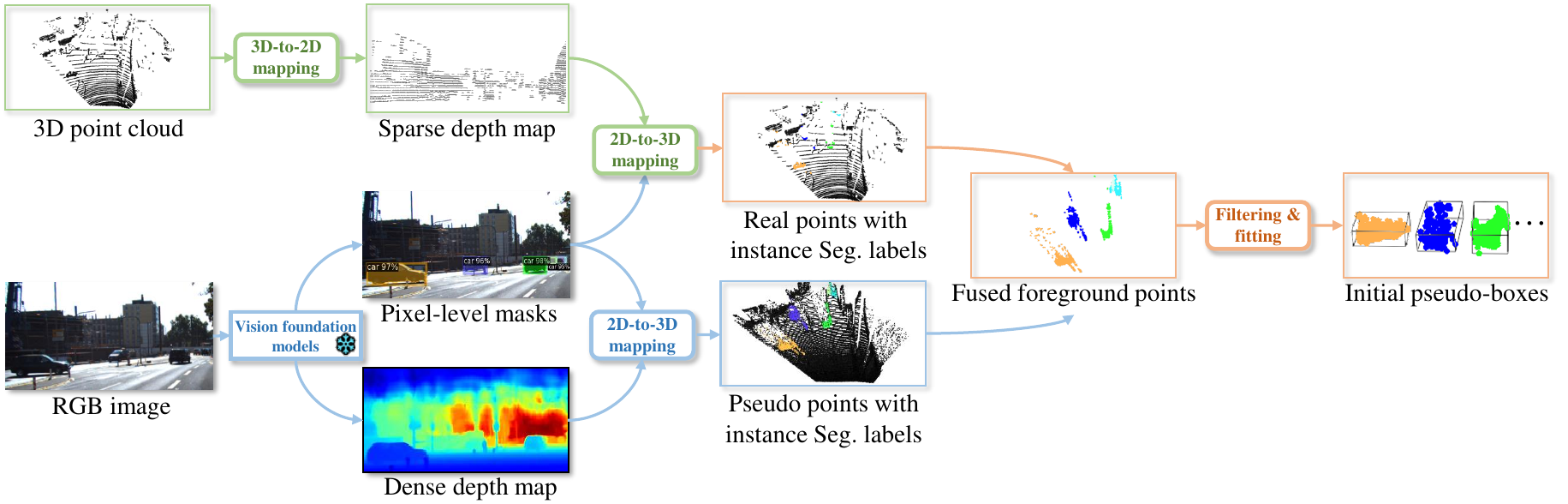}
       \caption{Illustration of the pseudo-box generation stage in DFU3D. We use vision foundation models to obtain instance masks and depth maps from RGB images, generating pseudo points to supplement real foreground points from point cloud while assigning class labels. After fusing both sets, we apply noise filtering to mitigate estimation errors from vision foundation models and generate initial 3D pseudo-boxes.}
       \Description{}
      \label{pipeline}
    \end{figure*}

\section{Related Work}

Since our unsupervised method is built upon typical 3D object detectors and uses self-evolution techniques for training, we briefly review recent works in the following fields: LiDAR-based 3D object detection, unsupervised 3D object detection, and self-evolution.

\subsection{LiDAR-based 3D Object Detection}
 Existing LiDAR-based 3D object detectors are generally divided into three categories based on their different data representations \cite{mao2022review}, including point-based, voxel-based, and point-voxel-based methods. Point-based methods \cite{yang2018ipod,yang2019std,shi2019pointrcnn,shi2020pointGNN,shi2020points} extract point features from the raw point cloud through a point-based backbone \cite{qi2017pointnet++}. Voxel-based methods \cite{yan2018second,lang2019pointpillars,deng2021voxel,yin2021center} extract voxel features from the voxelized points through the 3D sparse convolution \cite{graham20183d}. Point-voxel-based methods \cite{shi2020pv,shi2022pv,hu2022pdv} combine the point-based and voxel-based methods to extract the point and voxel features, respectively.  
 The above methods achieve good performance when being trained under the target scenes in a fully supervised way, but they are hard to generalize to new scenes where manual annotations are usually unavailable. In contrast, our method uses them as base detectors and investigates an effective unsupervised training mechanism.

\subsection{Unsupervised 3D Object Detection}
\paragraph{Single-modal methods.}
 Recently, single-modal methods for unsupervised 3D object detection aim to generate pseudo-boxes based on point cloud, and then train a typical base detector. You \textit{et al} \cite{you2022learning} apply the HDBSCAN algorithm \cite{mcinnes2017hdbscan} on selected mobile points to discover moving 3D objects and then use the estimated 3D bounding boxes as pseudo-boxes for training; Zhang \textit{et al} \cite{zhang2023towards} also discover moving objects based on point cloud sequences, but they employ shape prior knowledge to obtain more accurate pseudo-boxes for near-range moving objects with dense points, and utilize self-evolution \cite{xie2020self} to refine pseudo-boxes for long-range moving objects with sparse points; Wu \textit{et al} \cite{wu2024commonsense} introduce the Completeness and Size Similarity (CSS) score as a metric for evaluating the initial pseudo-boxes, and employ the common-sense size heuristic to refine pseudo-boxes that receive poor CSS scores.
 Although these works can effectively distinguish foreground objects, they suffer from point cloud sparsity, which limits the completeness of object representations, and the lack of category information from visual cues, making it difficult to generate accurate and reliable pseudo-boxes.

\paragraph{Multi-modal methods.}
Currently, multi-modal unsupervised methods mainly use class information or pseudo-boxes obtained from the image modality to compensate for the pseudo-boxes obtained solely from the point cloud modality. Najibi \textit{et al} \cite{najibi2023unsupervised} employ a vision foundation model on the image modality to estimate the categories of detection results during inference. Similarly, Lentsch \textit{et al} \cite{lentsch2024union} use a vision foundation model to determine the categories of pseudo-boxes generated from the LiDAR modality. Zhang \textit{et al} \cite{zhang2024approaching} directly leverage a vision foundation model to generate pseudo-boxes from the image modality at long ranges, complementing those obtained from the LiDAR modality. Although these label-level multi-modal fusion methods effectively compensate for the limitations of single-modal pseudo-boxes by incorporating category information or pseudo-boxes from the image modality, the quality of the pseudo-boxes remains dependent on the sparse point cloud. In contrast, our method adopts a data-level fusion method by integrating pseudo points from images with the original point cloud, enabling the generation of high-quality pseudo-boxes based on a denser representation.
 
\subsection{Self-Evolution}
 Self-evolution is proven to be effective in improving the quality of pseudo-boxes for semi-supervised learning \cite{xie2020self}, transfer learning \cite{phoo2020self,tian2020rethinking,ghiasi2021multi}, and domain adaptation \cite{zhang2019category,zou2019confidence,xie2020self}. Similarly, it has also been used by unsupervised 3D object detection methods \cite{you2022learning,zhang2023towards} to refine pseudo-boxes and retrain the detector. However, the limitations are that, significant computational resources are required to repeatedly train the detector based on the sparse point cloud over dozens of rounds for the methods  \cite{you2022learning,zhang2023towards}. Similarly, the approach presented by \cite{wu2024commonsense} demands substantial computational resources to train multiple detectors in a dense-sparse alignment manner. Different from them, we propose a dynamic pseudo-box evolution paradigm to refine the pseudo-boxes by locating local convergence points under the dense representation, leading to more reliable pseudo-boxes.

\section{Methodology}

In this section, we first give an overview of our method, and then provide a detailed introduction to each component.

\subsection{Overview}
We propose a data-level LiDAR-camera fusion framework for unsupervised 3D object detection, which fully integrates image and LiDAR information to generate high-quality class-aware pseudo-boxes. The pipeline consists of two key modules: data-level fusion based initial pseudo-box generation (see Sec. \ref{DLF}) and dynamic self-evolution (see Sec. \ref{DSE}).

Specifically, we first obtain instance segmentation masks using a vision foundation model and perform bi-directional mapping between the image and LiDAR modalities. Each real point is projected onto the 2D image space to acquire its class label and instance ID, while each 2D RGB pixel is back-projected onto 3D using a dense depth map estimated from a vision foundation model, generating pseudo points to supplement sparse real points. Then, we fuse real and pseudo foreground points, and introduce a local and global filtering method to suppress noise introduced by vision foundation models. Local radius filtering removes noise from depth estimation errors, while global statistical filtering eliminates outliers from segmentation errors. These ensure a more reliable and denser points, from which initial pseudo-boxes are generated via shape fitting.

Subsequently, we refine pseudo-boxes through a data-level fusion based dynamic update strategy. The pseudo foreground points from historical pseudo-boxes are merged with the real points to train the base detector, and pseudo-boxes are dynamically updated at the local convergence points, leading to more accurate pseudo-boxes.

    \begin{figure}[t]
      \centering
       \includegraphics[width=1.0 \linewidth]{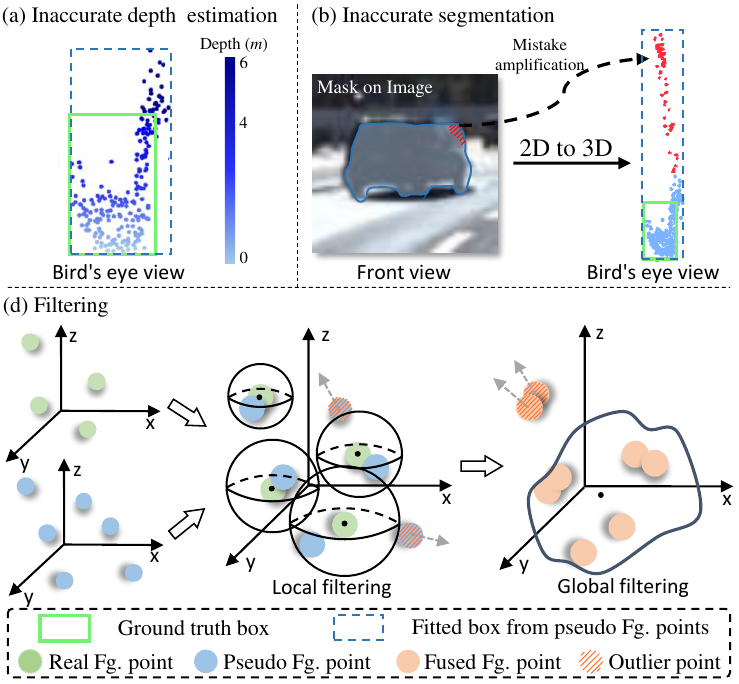}
       \caption{Visualization of depth and segmentation errors, along with our local and global filtering method. ``Fg." denotes foreground.
       }
       \Description{}
      \label{filtering}
    \end{figure}

% (a) Depth estimation errors cause pseudo points to deviate, distorting object geometry. (b) Segmentation errors misclassify points, introducing spatial outliers. (c) Our filtering method mitigates these errors via a local and global filtering, ensuring reliable fusion.

\subsection{Data-Level Fusion Based Initial Pseudo-Box Generation} \label{DLF}

\paragraph{Multi-modal bi-directional fusion.} \label{MBF} 
We introduce 2D RGB images as the auxiliary modality to obtain pixel-level instance segmentation labels via SEEM \cite{zou2023segmentSEEM} and dense depth maps via DepthAnything \cite{yang2024depthanything}. In this way, each pixel on the 2D RGB images obtains class and instance labels as well as a depth value. For real points from the LiDAR point cloud, we simply map them back to the 2D image space given camera parameters, and then pick the class and instance labels from corresponding pixels on the instance segmentation map. We generate pseudo points by mapping each pixel on 2D images to the 3D space using the estimated depth value, while each pseudo point takes along the class and instance label with it as it originates from the 2D image. We then merge real and pseudo points into a unified, high-density point set and filter out background points, retaining only foreground points. 
Since we deal with each object instance separately below, we omit the index of objects for simplicity and denote the real foreground point set of an arbitrary object as $R = \left \{r_1, r_2, \cdots \right \}$, where $r_i$ is one real point. Similarly, we denote the pseudo foreground point set of the same object as $V = \left \{v_1, v_2, \cdots \right \}$, where $v_i$ is one pseudo point.

 \paragraph{Local filtering for depth estimation noise.}
 Vision foundation models provide valuable semantic and depth information, but their estimation errors introduce a significant source of noise that degrades pseudo-box quality. In particular, depth estimation errors not only introduce substantial noise but also systematically distort object structures, leading to significant deviations in pseudo-box generation (see Figure \ref{filtering} (a)). To address this issue, we introduce a depth-aware ball query to filter out unreliable pseudo foreground points based on their Euclidean distance to the nearest real foreground point. Points exceeding an adaptive threshold are discarded, with different radius sizes applied for near and distant objects. A smaller radius is used in dense regions to remove redundant pseudo points, while a larger radius is applied in sparse regions to retain more pseudo foreground points and enrich geometric structure. This adaptive filtering ensures a more reliable and consistent fused foreground point set for subsequent processing. After this filtering step, the refined fused foreground point set for an object is defined as:
    \begin{equation}
      F = \left \{ v_j | E(r_i,v_j) \leq \lambda \cdot E(r_i, o) \right \} \cup R,
    \label{Fusion}
    \end{equation}
\noindent where $E(\cdot)$ is the Euclidean distance; $\lambda$ is a constant factor; $o$ is the origin point; $r_i$ and $v_j$ represent one real and one pseudo foreground point; and $R$ denotes the set of all real foreground points.

\paragraph{Global filtering for segmentation estimation noise.}
Apart from depth estimation errors, segmentation errors in 2D images also pose a critical challenge, as they can be amplified in 3D space, causing severe misalignment in object structure (see Figure \ref{filtering} (b)). For example, two adjacent pixels in the 2D segmentation mask may correspond to points that are meters apart in 3D due to depth misalignment, leading to misplaced outliers in the fused representation. To mitigate this, we introduce global statistical filtering, which removes outliers by analyzing the spatial consistency of each object. Specifically, we compute the mean distance of each point to its neighbors and eliminate those that significantly deviate from the distribution. The refined fused point set $U$ is defined as:
    \begin{equation}
      U = \left \{ f_m | E(f_m, f_n) \leq \Phi_m  \right \},
    \label{Filter}
    \end{equation}
 \noindent where $E(\cdot)$ is the Euclidean distance; $f_m$ and $f_n$ are two distinct points in the fused foreground point set $F$ defined in Equation (\ref{Fusion}); $\Phi_m$ is the threshold of statistical filtering \cite{anderson2012optimal}. Please see the supplementary material for the detailed explanation for $\Phi$.

\begin{figure}[t]
      \centering
       \includegraphics[width=1.0 \linewidth]{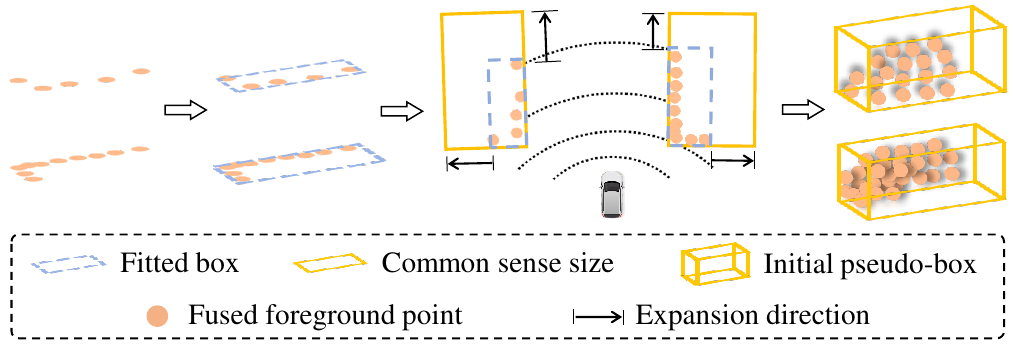}
       \caption{Illustration of the shape fitting process. It involves three steps: fitting 2D boxes in BEV, adjusting 2D box sizes based on common sense constraints, and lifting the refined 2D boxes to 3D.}
       \Description{}
      \label{box}
    \end{figure}

\paragraph{Pseudo-box generation.}
 Based on the cleaned foreground points, we fit them to obtain pseudo-boxes. As illustrated in Figure \ref{box}, we first apply the off-the-shelf shape fitting \cite{zhang2017efficient} to generate 2D BEV boxes. However, due to occlusion and missing signals \cite{xu2022behind}, some boxes deviate from reasonable object sizes. To correct this, we adjust boxes based on common sense sizes \cite{lentsch2024union} to correct unrealistically small boxes while maintaining shape consistency. Finally, we lift the adjusted 2D boxes to 3D using the topmost point in $U$ as the height reference, generating initial pseudo-boxes.
 
\subsection{Data-Level Fusion Based Dynamic Self-Evolution} \label{DSE}
After generating the initial pseudo-boxes, self-evolution can further refine them. However, existing methods operate in sparse point cloud space, limiting the model learning and degrading pseudo-box quality. Additionally, fixed-step self-evolution often leads to performance stagnation, resulting in redundant computations and inefficiency. To address these issues, we introduce \textit{Real Point Densification} to enhance point density for better feature learning and \textit{Dynamic Self-Evolution} to adaptively refine pseudo-boxes to improve training efficiency.

\paragraph{Real point densification.}
To densify the real points, we obtain dense pseudo points for foreground objects based on initial pseudo-boxes. Specifically, for each pseudo-box, we densely sample points within its enclosed region using depth information estimated from the image modality. These pseudo points serve to complement the sparse real points, particularly for distant objects where point cloud coverage is limited. The obtained pseudo foreground points are then merged with the real points, forming a denser and more informative points. These denser fused points alleviate the limitations of sparse real points in self-evolution, allowing the detector to learn more informative features, ultimately enhancing detection performance at inference.

\paragraph{Dynamic self-evolution.}
To further improve training efficiency, we then continuously monitor the loss values and establish a checkpoint once the loss stabilizes over a certain period, indicating that training has converged under the current pseudo-boxes. The stopping condition is defined as:
    \begin{equation}
    |t_{e} - t_{e-1}| \leq \psi pe^{-p},
    \label{condation}
    \end{equation}
\noindent where $t_e$ is the variance of the loss difference of previous epochs; $\psi$ is an exponential decay factor, and $p$ is the number of evolution phases. Once the loss satisfies this condition, we use the trained model to generate an updated set of pseudo-boxes. A newly detected box is retained only if its 3D intersection over union with any existing pseudo-box of the same class is lower than the threshold $v$ of the base detector; meanwhile, any previous pseudo-box that overlaps with the new one is discarded. This process refines the pseudo-boxes by selectively incorporating high-confidence detections. Subsequently, we continue training the model using the updated pseudo-boxes and the dense points. These steps are dynamically repeated within a single training round, progressively improving pseudo-box quality.

    % ---------------------------------------------------------------------------------------%
    \begin{algorithm}[t]
    % \begin{algorithm}[t]\scriptsize
        \caption{Data-level fusion based dynamic self-evolution}
        \label{algorithm}
        \raggedright
        \footnotesize
        \textbf{Input}: pseudo-boxes from phase ${p-1}$: $ \{ L^{p-1} \}$ ; 
        real points: $C_r$ ;
        pseudo points: $C_d$ ;
        threshold of 3D IoU: $\Psi$ ; 
        maximum epoch: $e_{max}$ \\
        % \textbf{Parameter}: Loss of epoch $e-1$: $l^{e-1}$ ; Loss of epoch $e$: $l^{e}$ \\
        \textbf{Output}: the refined pseudo-boxes after $p$: $\{ L^p \}$ \\
        \begin{algorithmic}[1]
        \STATE $p$ $\leftarrow$ $\mathrm{1}$ \\
        \FOR {$e \leftarrow 1$ \TO $e_{max}$}
        \IF {$|t_{e} - t_{e-1}| \leq \psi pe^{-p}$}
        \STATE // \textbf{\textit{Real point densification}}
        \STATE $V$ $\leftarrow$ Crop$(C_d, \{ L^{p-1} \})$ // Obtain pseudo foreground points
        \STATE $C_d$ $\leftarrow$ $C_r \cup V$ // Obtain dense points
        \STATE // \textbf{\textit{Dynamic self-evloution}}
        \STATE $\left\{ L^{p} \right\}$ $\leftarrow$ TestModel($C_d$) // Test the detector
        \STATE $I$ $\leftarrow$ $\mathrm{3D}$ $\mathrm{IoU}$ $\left( \{ L^{p-1} \}, \left\{ L^p \right\} \right)$  // Calculate 3D IoU matrix $p$
        \STATE $I_{max}$, $idx$ $\leftarrow$ $\mathrm{MaxAlongJ}$ ($I$) // Calculate the largest 3D IoU and corresponding index
        \STATE $I_{mask}$ $\leftarrow$ $I_{max}$ \textless \hspace{0.1cm} $v$
        \STATE $\left\{ L_{add}^p \right\}$ $\leftarrow$ $\mathrm{Select}$ $\left( \left\{ L^p \right\}, I_{mask} \right)$ // Add pseudo-boxes from phase $p$
        \STATE $\left\{ L_{res}^p \right\}$ $\leftarrow$ $\mathrm{Select}$ $\left( \{ L^{p-1} \}, idx \right)$ // Reserve pseudo-boxes from phase $p-1$
        \STATE $\left\{ L^p \right\}$ $\leftarrow$ $\mathrm{Concatenate}$ $\left( \{ L_{add}^p \}, \{L_{res}^p \} \right)$
        \STATE $p$ $\leftarrow$ $p + 1$
        \ENDIF
        \ENDFOR
        \STATE \textbf{return} $\left\{ L^p \right\}$
        \normalsize
        \end{algorithmic}
    \end{algorithm}
% ---------------------------------------------------------------------------------------%

    \begin{table*}[t]
        \centering
        \caption{Class-agnostic object detection on the nuScenes validation set. Results are obtained by training PointRCNN \cite{shi2019pointrcnn} with the generated pseudo-boxes. L and C are abbreviations for LiDAR and camera, respectively. Numbers are AP$_{BEV}$ / AP$_{3D}$.}
        \setlength{\tabcolsep}{2.0mm}{
        \scalebox{1.0}{
        \begin{tabular}{c|c|c|c c c|c} 
        \hline
        Methods & Labels & Modality & 0-30$m$ & 30-50$m$ & 50-80$m$ & 0-80$m$ \\
        \hline
        Supervised & 1\% & - & 9.7 / 7.7 & 3.5 / 2.0 & 0.7 / 0.6 & 4.2 / 3.4 \\
        Supervised & 10\% & - & 25.2 / 21.4 & 6.3 / 5.0 & 2.7 / 1.8 & 10.9 / 8.5 \\
        Supervised & 100\% & - & 39.8 / 34.5 & 12.9 / 10.0 & 4.4 / 2.9 & 22.2 / 18.2 \\
        \hline
        MODEST \cite{you2022learning} & 0 & L & 24.8 / 17.1 & 5.5 / 1.4 & 1.5 / 0.3 & 11.8 / 6.6 \\
        OYSTER \cite{zhang2023towards} & 0 & L & 26.6 / 19.3 & 4.4 / 1.8 & 1.7 / 0.4 & 12.7 / 8.0 \\
        LiSe \cite{zhang2024approaching} & 0 & L+C & 35.0 / 24.0 & 11.4 / 4.4 & 4.8 / 1.3 & 19.8 / 11.4 \\
        \rowcolor{gray!20}
        DFU3D (ours) & 0 & L+C & 37.2 / 29.4 & 14.0 / 9.1 & 4.8 / 2.8 & \textbf{20.7} / \textbf{15.0} \\
        \hline
        \end{tabular}
        }
        }
        \label{Class-agnostic experiment}
    \end{table*}

    \begin{table*}[t]
        \centering
        \caption{Class-aware object detection on the nuScenes validation set. Results are obtained by training CenterPoint \cite{yin2021center} with the generated pseudo-bounding boxes.  *Results are taken from \cite{lentsch2024union}. Per-class values are AP$_{3D}$; the last column shows the mAP$_{3D}$.}
        \setlength{\tabcolsep}{2.5mm}{
        \scalebox{1.0}{
        \begin{tabular}{c|c|c|c c c|c} 
        \hline
        Methods & Labels & Modality & Vehicle & Pedestrian & Cyclist & All \\
        \hline
        Supervised & 1\% & - & 39.3 & 31.8 & 1.8 & 24.3 \\
        Supervised & 10\% & - & 65.3 & 57.6 & 14.9 & 45.9 \\
        Supervised & 100\% & - & 75.4 & 74.2 & 45.9 & 65.2 \\
        \hline
        HDBSCAN* \cite{mcinnes2017hdbscan} & 0 & L & 14.1 & 0.4 & 0.0 & 4.9 \\
        UNION \cite{lentsch2024union} & 0 & L+C & 31.0 & 44.2 & 0.0 & 25.1 \\
        \rowcolor{gray!20}
        DFU3D (ours) & 0 & L+C & 32.3 & 37.7 & 15.3 & \textbf{28.4} \\
        \hline
        \end{tabular}
        }
        }
        \label{Class-aware object detection}
    \end{table*}

\section{Experiments}
In this section, we will first introduce the experimental setup, followed by the implementation details. Then, we will report the main results. In the end, we will show the ablation studies of our method.

\subsection{Experimental Setup}

\paragraph{Datasets.} We evaluate our method on the nuScenes datasets \cite{caesar2020nuscenes}. It consists of 700, 150, and 150 scene sequences for training, validation, and testing, respectively. Each sequence has 40 frames, where each frame contains six surrounding images covering a 360-degree field of view. It offers calibration matrices that facilitate accurate projection of 3D points onto 2D pixels, and contains 10 object categories that are commonly encountered within autonomous driving. 

\paragraph{Object classes.} In previous unsupervised 3D object detection methods, pseudo-boxes can be either category-agnostic or category-aware. Since our method generates category-aware pseudo-boxes, we adopt different evaluation strategies for comparison. For comparison with category-agnostic methods, we merge all detected objects into a single object category for evaluation. For comparison with category-aware methods, we follow the categorization scheme used in UNION \cite{lentsch2024union}, grouping objects into three major categories: (1) vehicle, including car, bus, construction vehicle, trailer, and truck classes; (2) cyclist, including bicycle and motorcycle classes; and (3) pedestrian.

\paragraph{Metrics.} For comparison with category-agnostic methods, we follow LiSe \cite{zhang2024approaching} to report evaluation results using average precision (AP) in both BEV (AP$_{BEV}$) and 3D space (AP$_{3D}$) at IoU = 0.25. The evaluation is also conducted across different object ranges: near (0–30$m$), middle (30–50$m$), far (50–80$m$), and full range (0–80$m$). For comparison with category-aware methods, we follow UNION \cite{lentsch2024union} to report evaluation results using AP using the standard nuScenes evaluation protocol. It is important to note that previous works do not provide speed and attribute information for pseudo-boxes, so the nuScenes detection score cannot be used for evaluation here.

\paragraph{Base detectors.} For a fair comparison, we follow LiSe \cite{zhang2024approaching} to use PointRCNN \cite{shi2019pointrcnn} as the base detector for category-agnostic experiments. For category-aware experiments, we adopt CenterPoint \cite{yin2021center} as the base detector, following UNION \cite{lentsch2024union}.

\subsection{Implementation Details}
We use the OpenPCDet framework \cite{od2020openpcdet} for all our experiments under 4 $\times$ NVIDIA 3090 GPUs. For PointRCNN, we follow LiSe to train the model for 80 epochs with a batch size of 8, and for CenterPoint, we follow UNION to train the model for 20 epochs with a batch size of 4. We adopt the default optimizer settings, learning rate schedule, and data augmentation strategies provided by each base detector. For the hyperparameters of DFU3D, we follow CenterPoint's configuration to set $v$ to 0.2 in the dynamic self-evolution module; $\lambda$ as a constant factor in Equation (\ref{Fusion}) is set to 0.01; $\psi$ as the exponential decay factor in Equation (\ref{condation}) is set to 0.1, which are mentioned in ablation studies in Section \ref{ablation}.

\subsection{Main Results}
\paragraph{Class-agnostic 3D object detection.}
We evaluate the performance of DFU3D for class-agnostic 3D object detection on the nuScenes validation set. In Table \ref{Class-agnostic experiment}, our DFU3D achieves 20.7\% AP$_{BEV}$ and 15.0\% AP$_{3D}$ in the 0-80$m$ range, surpassing LiSe \cite{zhang2024approaching} by 0.9 and 3.6 $pp$, demonstrating the effectiveness of our data-level fusion strategy in improving pseudo-box quality. Compared to fully supervised methods, DFU3D significantly narrows the performance gap. In the 0-80$m$ range, our method achieves comparable performance (22.2\% AP$_{BEV}$ \textit{vs.} 20.7\% AP$_{BEV}$) to fully supervised learning, which utilizes 100\% labeled data. Moreover, in the 30-50$m$ range, DFU3D even surpasses the fully supervised counterpart, achieving 14.0\% AP$_{BEV}$ compared to 12.9\% AP$_{BEV}$. These results indicate that our data-level fusion method outperforms state-of-the-art unsupervised methods, and the detector trained with our approach achieves performance comparable to a fully supervised model, even surpassing it in mid-range detection. These results demonstrate the effectiveness of our method.

\paragraph{Class-aware 3D object detection.}
For more challenging class-aware 3D object detection, as shown in Table \ref{Class-aware object detection}, DFU3D achieves 28.4\% mAP, surpassing the state-of-the-art UNION by 3.3 $pp$, further demonstrating the effectiveness of our method. Notably, our method successfully generates high-quality pseudo-boxes that enable the detector to achieve 15.3\% mAP for the cyclist category, for which UNION completely fails (0.0 mAP). Furthermore, the detector trained by DFU3D is comparable to a fully supervised model trained with 10\% labeled data (14.9\% mAP) on the cyclist category, underscoring the strength of our data-level fusion strategy. These results suggest that our approach not only improves overall detection performance but also enhances recognition for rare categories, making it a more practical solution for real-world applications.

Although our method performs well on rare categories like cyclists, a performance gap remains in the pedestrian category compared to UNION. We attribute this to the limitations of the vision foundation model, which tends to generate noisy pseudo points. As illustrated in Figure \ref{lambda} ($\lambda$=1), many noisy points appear around pedestrians. Although our filtering module alleviates most errors, some noise remains near true pedestrian points, leading to suboptimal box orientation and size, and thus lower detection accuracy.

\subsection{Ablation Studies and Analyses} \label{ablation}

\paragraph{Effectiveness of multi-modal bi-directional fusion.} 
To validate the effectiveness of our bi-directional fusion method at the data level, we conduct ablation experiments using three variants: (1) $Q_1$: real foreground points only, (2) $Q_2$: virtual foreground points only, and (3) $Q_3$: a fusion of both. The results are shown in Table \ref{bi-directional ablation}. We observe that the detector trained with fused foreground points achieves better performance compared to using only real points or virtual points, demonstrating the advantage of bi-directional fusion. We attribute this improvement to the higher point density achieved by fusing real and virtual points, which helps alleviate the sparsity problem in regions with limited real points. These denser foreground points are particularly beneficial for small objects like cyclists and pedestrians, which are more sensitive to point sparsity, leading to more accurate pseudo-box generation.

\begin{figure}[t]
      \centering
       \includegraphics[width=1.0 \linewidth]{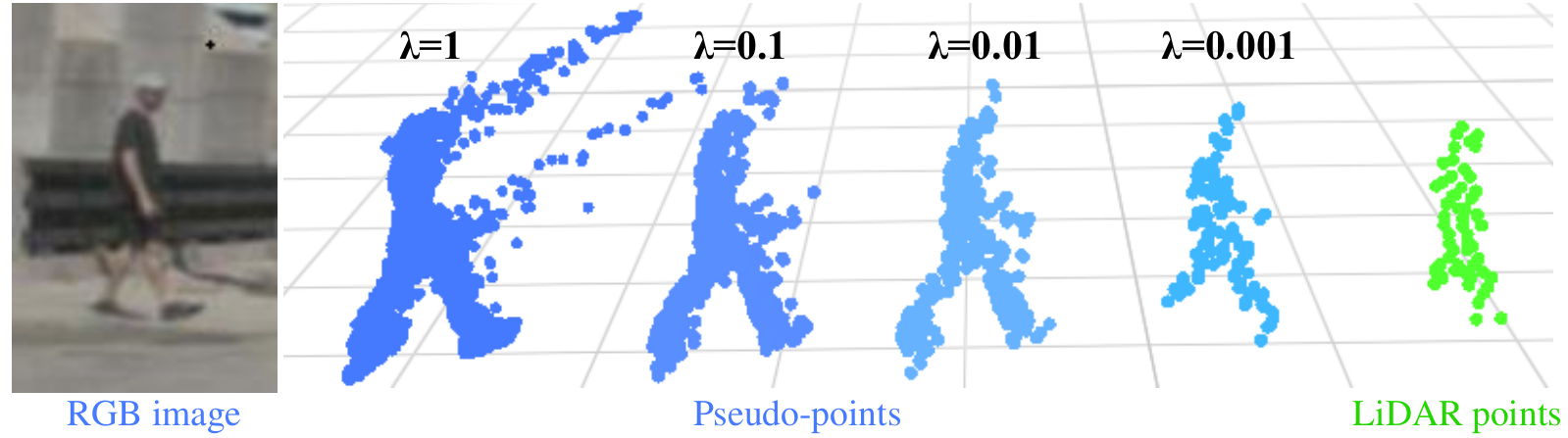}
       \caption{Illustration of the selection of $\lambda$ in Equation (\ref{Fusion}) in the local filtering module.}
       \Description{}
      \label{lambda}
    \end{figure}

\paragraph{Effectiveness of local and global filtering.} 
To evaluate the effectiveness of local and global filtering strategies in initial pseudo-box generation, we conduct ablation experiments with three configurations: (1) $Q_4$: applying only local filtering, (2) $Q_5$: applying only global filtering, and (3) $Q_6$: combining both local and global filtering. We observe that filtering out noise points—either locally or globally—significantly improves the quality of pseudo-boxes. Moreover, the combination of both filtering strategies ($Q_6$) yields the most substantial performance gain. Among all categories, we observe the most remarkable improvement with a 25.4 $pp$ increase from $Q_3$ to $Q_6$ in pedestrian detection. This can be attributed to the fact that pedestrians are relatively small in size and more susceptible to the influence of surrounding noise. As visualized in Figure \ref{lambda}, when the filtering strength is weak ($\lambda=1$), the outlier points significantly change the pedestrian shape. By effectively eliminating these noise points, our filtering strategy significantly enhances the pseudo-box quality. These results demonstrate the necessity and effectiveness of our proposed filtering approach.

    \begin{table}[t]
        \centering
        \caption{Ablation study of each component of pseudo-box generation. RP: real point; PP: pseudo point; LF: local filtering; GF: global filtering} 
        \setlength{\tabcolsep}{1.5mm}{
        \scalebox{1.0}{
        \begin{tabular}{c| c c c c |c c c|c} 
        \hline
        ID & RP & PP & LF & GF & Vehicle & Pedestrian & Cyclist & All \\
        \hline
        $Q_1$ & \Checkmark & & & & 15.8 & 8.1 & 1.5 & 8.5 \\
        $Q_2$ & & \Checkmark & & & 11.8 & 8.9 & 1.7 & 7.5 \\
        $Q_3$ & \Checkmark & \Checkmark & & & 15.9 & 9.3 & 3.5 & 9.6 \\
        $Q_4$ & \Checkmark & \Checkmark & \Checkmark & & 26.1 & 20.5 & 9.8 & 18.8 \\
        $Q_5$ & \Checkmark & \Checkmark & & \Checkmark & 27.3 & 15.5 & 10.3 & 17.7 \\
        $Q_6$ & \Checkmark & \Checkmark & \Checkmark & \Checkmark & \textbf{28.0} & \textbf{34.7} & \textbf{10.9} & \textbf{24.5} \\
        \hline
        \end{tabular}
        }
        }
        \label{bi-directional ablation}
    \end{table}

    \begin{table}[t]
        \centering
        \caption{Ablation study of data-level fusion based dynamic self-evolution. ``Fixed-$X$" and ``Dynamic-$X$" refer to two update strategies with $X$ round(s) of pseudo-box self-evolution.}
        \setlength{\tabcolsep}{1.5mm}{
        \scalebox{1.0}{
        \begin{tabular}{c|c c|c c c|c} 
        \hline
        \multirow{2}{0.3cm}{ID} & Pseudo & Update & \multirow{2}{0.8cm}{Vehicle} & \multirow{2}{1.2cm}{Pedestrian} & \multirow{2}{0.8cm}{Cyclist} & \multirow{2}{0.4cm}{All} \\
         & Points & Strategies & & & & \\
        \hline
        $Q_7$ & \ding{55} & Fixed-1 & 28.7 & 34.8 & 11.5 & 25.0 \\
        $Q_8$ & \Checkmark & Fixed-1 & 30.3 & 35.9 & 12.5 & 26.2 \\
        $Q_9$ & \Checkmark & Dynamic-1 & 32.3 & \textbf{37.7} & 15.3 & 28.4 \\
        $Q_{10}$ & \Checkmark & Dynamic-2 & 34.0 & 37.4 & 14.7 & 28.7 \\
        $Q_{11}$ & \Checkmark & Dynamic-3 & \textbf{34.2} & 37.3 & \textbf{15.6} & \textbf{29.0} \\
        $Q_{12}$ & \Checkmark & Dynamic-4 & 33.9 & 37.5 & 15.0 & 28.8 \\
        \hline
        \end{tabular}
        }
    }
    \label{self-evolution ablation}
    \end{table}

\begin{figure}[t]
      \centering
      \begin{subfigure}[t]{0.49 \linewidth}
        \includegraphics[width=1 \linewidth]{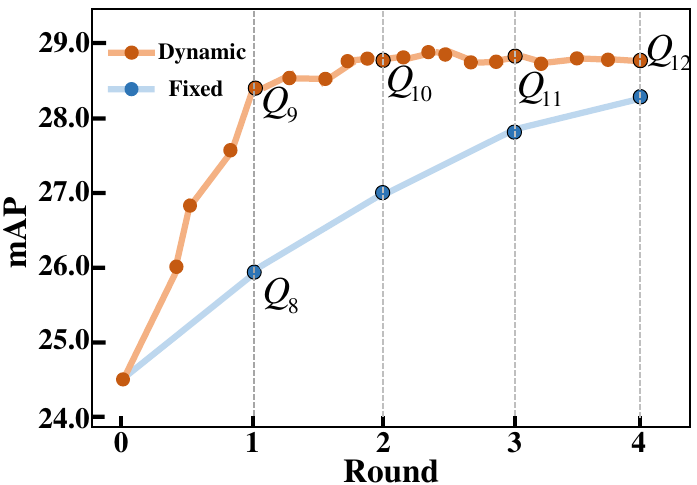}
        \caption{Dynamic and fixed-step update strategy}
        \label{Q7toQ12}
      \end{subfigure}
      \hfill
      \begin{subfigure}[t]{0.49 \linewidth}
        \includegraphics[width=1 \linewidth]{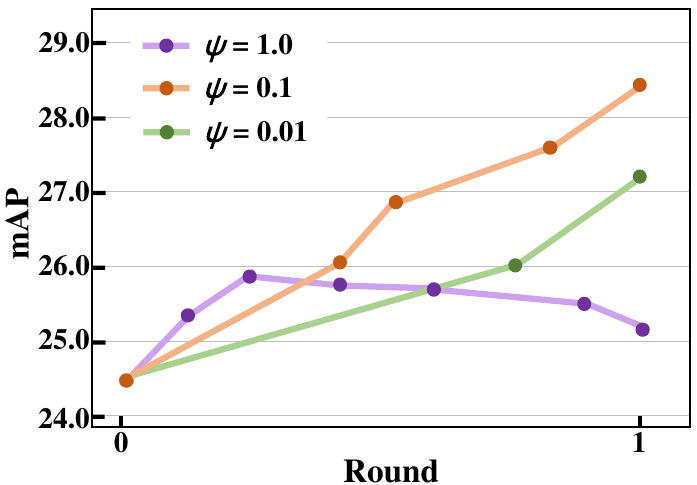}
        \caption{Impact of exponential decay factor $\psi$ on updates}
        \label{Q7toQ10}
      \end{subfigure}
      \caption{Performance trends under different pseudo-box update strategies. The colored points of the lines represent the updating position of the pseudo-boxes.}
      \label{ablation_psi}
    \end{figure}

\paragraph{Effectiveness of data-level fusion based dynamic self-evolution.}
As shown in Table \ref{self-evolution ablation}, we observe that, compared to the previous self-evolution method ($Q_7$), enhancing the density of foreground points with pseudo points ($Q_8$) consistently improves the quality of pseudo-boxes across all categories. This demonstrates the effectiveness of our point densification strategy in compensating for sparse real points. Furthermore, when adopting our dynamic update strategy, as illustrated in Figure \ref{Q7toQ12}, we observe that under the same number of training rounds, our method ($Q_9$) produces higher-quality pseudo-boxes than the fixed-step strategy ($Q_8$). Moreover, given the same number of pseudo-box updates, our dynamic strategy achieves comparable performance to the fixed-step method using four update rounds, showcasing both the efficiency and effectiveness of our dynamic self-evolution mechanism. However, when training with additional rounds, the performance only marginally improves after the second round ($Q_{10}$), and eventually saturates as observed in longer training ($Q_{11}$, $Q_{12}$). Taking efficiency into account, we adopt one update round as the default setting in our method.

    \begin{table}[t]
        \centering
        \caption{Ablation study of $\lambda$ in Equation (\ref{Fusion}) in the local filtering module.}
        \setlength{\tabcolsep}{1.7mm}{
        \scalebox{0.97}{
        \begin{tabular}{c|c c c|c} 
        \hline
         $\lambda$ & Vehicle & Pedestrian & Cyclist & All \\
        \hline
        1 & 26.3 & 14.7 & 9.4 & 16.8 \\
        0.1 & 27.4 & 31.5 & 9.5 & 22.8 \\
        0.01 & 28.0 & \textbf{34.7} & \textbf{10.9} & \textbf{24.5} \\
        0.001 & \textbf{28.4} & 27.7 & 9.2 & 21.8 \\
        \hline
        \end{tabular}
        }
    }
    \label{lambda ablation}
    \end{table}

    \begin{table}[t]
        \centering
        \caption{Ablation study of $\psi$ in Equation (\ref{condation}) in the dynamic self-evolution.}
        \setlength{\tabcolsep}{1.7mm}{
        \scalebox{0.97}{
        \begin{tabular}{c |c c c|c} 
        \hline
        $\psi$ & Vehicle & Pedestrian & Cyclist & All \\
        \hline
        1 & 28.7 & 33.5 & 13 & 25.1 \\
        0.1 & \textbf{32.3} & \textbf{37.7} & \textbf{15.3} & \textbf{28.4} \\
        0.01 & 30.9 & 36.8 & 13.9 & 27.2 \\
        \hline
        \end{tabular}
        }
    }
    \label{psi ablation}
    \end{table}

    \begin{figure*}[t]
      \centering
       \includegraphics[width=1.0 \linewidth]{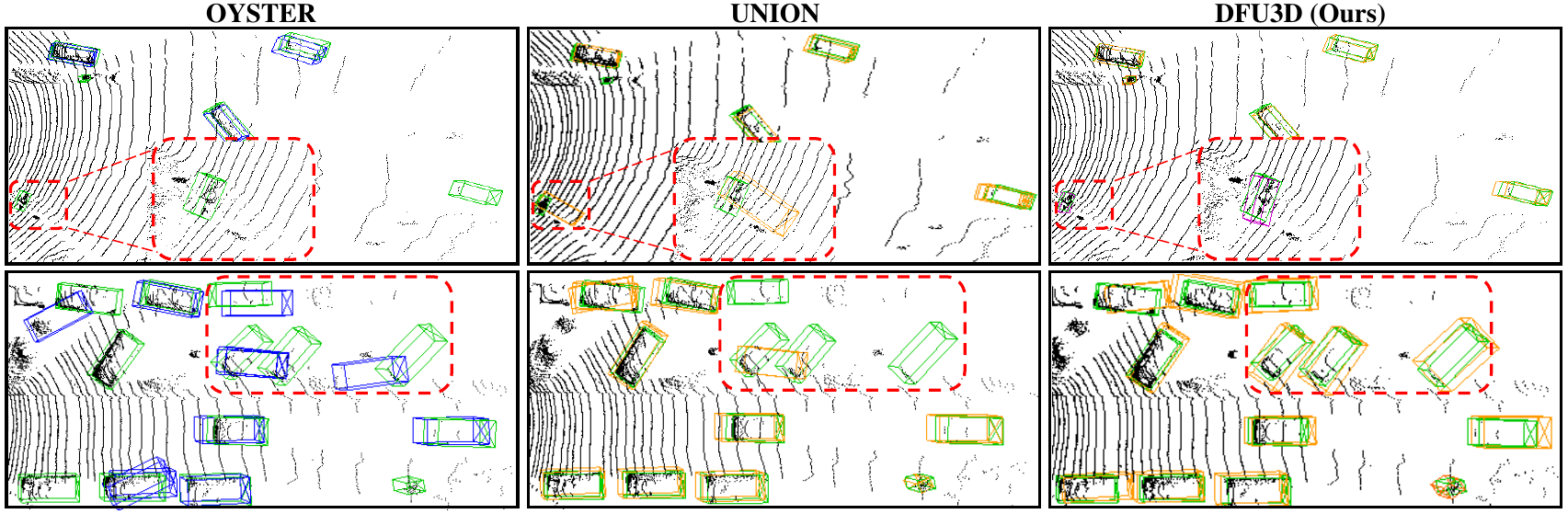}
       \caption{Comparison of detection bounding boxes among OYSTER \cite{zhang2023towards}, UNION \cite{wu2024commonsense}, DFU3D (ours). \textcolor{green}{Green} boxes are the ground truth boxes, \textcolor{blue}{blue} boxes represent the class-agnostic predicted results, \textcolor[rgb]{1, 0.6, 0.0}{orange} boxes represent the predicted results for the vehicle category, \textcolor[rgb]{0.8, 0.0, 0.8}{purple} boxes represent the predicted results for the cyclist category, and \textcolor{red}{red} boxes represent the predicted results for the pedestrian category. \textcolor{red}{Red} rectangles highlight the main differences in predictions.}
      \label{visualization}
    \end{figure*}

\paragraph{Ablation studies of $\lambda$ in Equation (\ref{Fusion}).}
We conduct ablation studies to investigate the impact of the filtering strength $\lambda$ in Equation (\ref{Fusion}), where local filtering plays a critical role, particularly for small objects like pedestrians (as previously indicated in Table \ref{bi-directional ablation}). We vary $\lambda$ across several magnitudes to examine its influence on the pseudo-box quality. As listed in \ref{lambda ablation}, the results show that $\lambda=0.01$ achieves the highest 34.7\% AP for pedestrians. As visualized in Figure \ref{lambda}, we analyze that weak filtering ($\lambda=1$) retains excessive noise from depth estimation errors, degrading box fitting accuracy, while overly aggressive filtering ($\lambda=0.001$) removes essential structural points, resulting in distorted object shapes. In contrast, moderate values such as $\lambda=0.1$ or $\lambda=0.01$ successfully eliminate most outliers and preserve the object geometry. Among these, $\lambda=0.01$ yields the best pseudo-box quality. These findings support our choice of $\lambda$ for optimal local filtering.
    
\paragraph{Ablation studies of $\psi$ in Equation (\ref{condation}).}
In Table \ref{psi ablation}, by testing $\psi$ across several magnitudes, we examine its effect on pseudo-box quality. The results show that $\psi=0.1$ achieves the best 28.4\% mAP. In Figure \ref{ablation_psi}, we analyze that a large value of $\psi$ ($\psi=1$) relaxes the convergence criterion, which may trigger pseudo-box generation before the base detector reaches a relatively stable state, thereby compromising the quality of subsequent pseudo-box updates. In contrast, a small value of $\psi$ ($\psi=0.01$) imposes a stricter condition, which slows down the update process and reduces training efficiency. Overall, these results validate our selection of $\psi$ as a balance between efficient convergence and reliable pseudo-box quality.

\paragraph{Ablation studies of vision foundation models.}
As shown in Table \ref{VFM ablation}, when we keep the segmentation model fixed as SEEM, and replace DepthAnything with the lightweight PENet \cite{hu2021penet}, the overall performance drops by 4.3 $pp$, despite obatining faster inference. Conversely, when we fix DepthAnything, and replace SEEM with the classical HTC\_ResNeXt101 \cite{chen2019hybrid}, the performance also degrades. These results indicate that high-quality depth estimation and accurate segmentation are critical for generating accurate pseudo-boxes. It is worth noting that our main paper contributions do not depend on specific vision foundation models. DFU3D is flexible and can incorporate more advanced depth or segmentation models as they become available.

\paragraph{Qualitative Analysis.} 
We visualize the detection bounding boxes in Figure \ref{visualization}. In the first row, DFU3D accurately identifies a nearby cyclist, which belongs to a rare category, within the red rectangle. In contrast, OYSTER fails to detect it, and UNION misclassifies it as a vehicle. In the second row, DFU3D produces more precise bounding boxes for occluded objects, while the other methods either fail to detect them or generate inaccurate boxes. These results further demonstrate our DFU3D not only enhances the recognition of rare categories, but also improves bounding box precision.

    \begin{table}[t]
        \centering
        \caption{Ablation study of vision foundation models in DFU3D. We fix either the segmentation or depth estimation model to analyze their individual impact. DepthAnything \cite{yang2024depthanything} and SEEM \cite{zou2023segmentSEEM} are used as default models in our main results.}
        \setlength{\tabcolsep}{1.4mm}{
        \scalebox{0.89}{
        \begin{tabular}{c|c|c c|c} 
        \hline
        \multicolumn{4}{c}{\textbf{(a) Fixed segmentation model: SEEM}} \\
        \hline
        Depth Estimation & Present at & Param. ($M$) & Runtime ($ms$) & All \\
        \hline
        DepthAnything \cite{yang2024depthanything} & CVPR'24 & 1281.2 & 258.5 & \textbf{24.5} \\
        PENet \cite{hu2021penet} & ICRA'21 & 502.9 & 19.5 & 20.2 \\
        \hline
        \hline
        \multicolumn{4}{c}{\textbf{(b) Fixed depth model: DepthAnything}} \\
        \hline
        Segmentation & Present at & Param. ($M$) & Runtime ($ms$) & All \\
        \hline
        SEEM \cite{zou2023segmentSEEM} & NeurIPS'23 & 627.8 & 183.3 & \textbf{24.5} \\
        HTC\_ResNeXt101 \cite{chen2019hybrid} & CVPR'19 & 546.4 & 187.7 & 23.0 \\
        \hline
        \end{tabular}
        }
        }
        \label{VFM ablation}
    \end{table}

\section{Conclusion}
In this paper, we propose data-level fusion for unsupervised 3D object detection. By introducing RGB images, we obtain class labels for both real and virtual points generated from the dense depth map; then we generate 3D pseudo-boxes by shape fitting on filtered foreground points. Moreover, our data-level fusion based self-evolution paradigm is more effective and efficient in dynamically refining pseudo-boxes under the dense fused points. Extensive experiments on the nuScenes dataset show that the model trained by our method outperforms previous unsupervised methods. We hope our work will inspire more research in the field of pseudo-box generation via data-level fusion for unsupervised 3D object detection.

\section*{Acknowledgment}
This research is supported by the National Natural Science Foundation of China (Grant No. 62322602), the Natural Science Foundation of Jiangsu Province, China (Grant No. BK20230033), and the National Natural Science Foundation of China (Grant No. 62172225).

\bibliographystyle{ACM-Reference-Format}
\balance
\bibliography{sample-base}

\end{document}